\documentclass[letterpaper]{article} 
\usepackage{aaai23}  
\usepackage{times}  
\usepackage{helvet}  
\usepackage{courier}  
\usepackage[hyphens]{url}  
\usepackage{graphicx} 
\usepackage{natbib}  
\usepackage{caption} 
\usepackage{algorithm}
\usepackage{algorithmic}
\usepackage{booktabs}
\usepackage{fancyhdr} 
\usepackage{newfloat}
\usepackage{listings}
\DeclareCaptionStyle{ruled}{labelfont=normalfont,labelsep=colon,strut=off} 
\floatstyle{ruled}
\newfloat{listing}{tb}{lst}{}
\floatname{listing}{Listing}
\title{ESL-SNNs: An Evolutionary Structure Learning Strategy for Spiking Neural Networks}
\author{
    Jiangrong Shen\textsuperscript{\rm1},
    Qi Xu\textsuperscript{\rm2}\textsuperscript{\rm*},
    Jian K. Liu\textsuperscript{\rm3},
    Yueming Wang\textsuperscript{\rm1},
    Gang Pan\textsuperscript{\rm1},
    Huajin Tang\textsuperscript{\rm1,\rm4}\thanks{Corresponding authors.}
}
\affiliations{
    \textsuperscript{\rm 1} The College of Computer Science and Technology, Zhejiang University, China \\
    \textsuperscript{\rm 2} School of Artificial Intelligence, Dalian University of Technology, China\\
    \textsuperscript{\rm 3} School of Computing, University of Leeds, UK \\ 
    \textsuperscript{\rm 4} Research Institute of Intelligent Computing, Zhejiang Lab, China \\ 
    jrshen@zju.edu.cn, xuqi@dlut.edu.cn, j.liu9@leeds.ac.uk, \{ymingwang, gpan, htang\}@zju.edu.cn 
    
}

\usepackage{bibentry}

\begin{document}

\maketitle


\begin{abstract}
Spiking neural networks (SNNs) have manifested remarkable advantages in power consumption and event-driven property during the inference process. To take full advantage of low power consumption and improve the efficiency of these models further, the pruning methods have been explored to find sparse SNNs without redundancy connections after training. However, parameter redundancy still hinders the efficiency of SNNs during training. In the human brain, the rewiring process of neural networks is highly dynamic, while synaptic connections maintain relatively sparse during brain development. Inspired by this, here we propose an efficient evolutionary structure learning (ESL) framework for SNNs, named ESL-SNNs, to implement the sparse SNN training from scratch. The pruning and regeneration of synaptic connections in SNNs evolve dynamically during learning, yet keep the structural sparsity at a certain level. As a result, the ESL-SNNs can search for optimal sparse connectivity by exploring all possible parameters across time. Our experiments show that the proposed ESL-SNNs framework is able to learn SNNs with sparse structures effectively while reducing the limited accuracy. The ESL-SNNs achieve merely $0.28\%$ accuracy loss with $10\%$ connection density on the DVS-Cifar10 dataset. Our work presents a brand-new approach for sparse training of SNNs from scratch with biologically plausible evolutionary mechanisms, closing the gap in the expressibility between sparse training and dense training. 
Hence, it has great potential for SNN lightweight training and inference with low power consumption and small memory usage.

\end{abstract}

\section{Introduction}

Cognitive functions of the brain stem from the complex neural networks composed of billions of neurons and trillions of synaptic connections between neurons. Synaptic connections form the physical basis for information communication within the brain. During the brain developmental process, synaptic connections rewire through the structural plasticity mechanism by forming new synapses and eliminating existing synapses \cite{de2017ultrastructural, barnes2010sensory, bennett2018rewiring}. Meanwhile, the rewiring process promotes the sparsity of synaptic connections and further facilitates the low-power consumption of neural systems of the brain. Inspired by the information processing and learning mechanism of the brain, Spiking neural networks (SNNs) have attracted increasing attention due to their highly biological plausibility and energy efficiency on neuromorphic chips. 
However, lacking structural plasticity makes SNNs training and inference suffer the bottleneck similar to traditional artificial neural networks, i.e., high power consumption and big memory usage due to parameter redundancy \cite{han2015learning}.

\begin{figure*}[t]
	\centering
	\includegraphics[width=2\columnwidth]{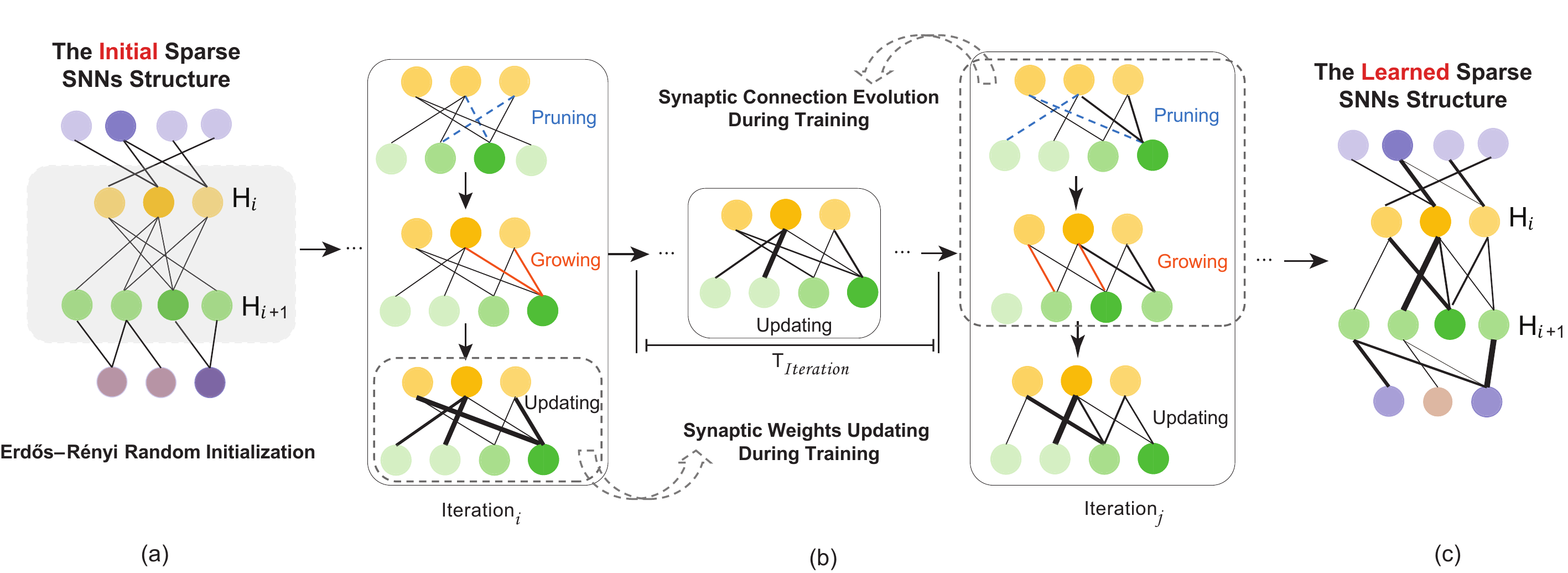}
	\caption{The evolution process of structure learning for ESL-SNNs. (a) The sparse SNNs are initialized by Erdös–Rényi random graph. Hence, each synaptic connection has a fixed probability of being present or absent, independent of other synaptic connections. (b) During the training process, the connection density maintains at a certain level to enable sparse training. Synaptic weights update in each iteration by the global learning rule, while the synaptic connection pruning and growth procedures proceed every $T_{iter}$ according to the pruning and growth rule, respectively. (c) The learned sparse structure is generated after the training process.}
	\label{fig_set}
\end{figure*}

Recently, some studies focus on learning the sparse structure of SNNs. To optimize network structure and connection weights, the synaptic sampling method is utilized by considering the spine motility of SNNs as Bayesian learning \cite{kappel2015network}.
Deep rewiring (Deep R) pruning algorithm \cite{bellec2018long} is proposed and applied on SpiNNaker 2 prototype chips \cite{liu2018memory}.
Gradient rewiring (Grad R) is proposed for connection learning \cite{chenpruning}, where the gradient is redefined to a new synaptic parameter to allow the connection competition between growing and pruning. The above studies have revealed structure refining capability in SNNs and indicated the high parameter redundancy in deep SNNs. 
\cite{chen2022state} facilitates weight optimization for pruning by modeling weight states, which employs the weights reparameterization method to find the final optimized sparse structure.
However, these models still could not implement stable sparse learning during the prolonged training process. The stable sparse learning emphasizes that the topology sparsity of SNNs is pursued starting with the network design phase, which leads to the substantial reduction in connections and, in turn, to memory and computation efficiency \cite{mocanu2018scalable}.

This paper implements the sparse training of SNNs from scratch by a sparse structure learning method inspired by biological neuronal networks. The proposed ESL-SNNs  not only employ the sparse structure during the inference process but also can be trained with the sparse structure, in which synaptic connections are pruned or regenerated dynamically during the training process (Figure \ref{fig_set}). The ESL-SNNs framework makes a step to close the gap between sparse training and dense training and tries to improve the expressibility of sparse training by the dynamic parameter exploration during the training process inspired by the rewiring mechanism in the brain, instead of inheriting weights from the dense and pre-trained model.
The contributions in this paper are summarized as follows:

\begin{itemize}

\item The sparse structure learning from scratch of SNNs is emphasized to reduce energy consumption and memory usage during both the training and inference processes. Significantly, the biologically plausible structure learning framework of ESL-SNNs is proposed to implement genuine sparse training, and it adapts to different kinds of SNNs architectures.

\item The proposed structure learning of SNNs can prune and regenerate synaptic connections dynamically during the training process, inspired by the rewiring mechanism in the brain. The connection rewiring in the ESL-SNNs promotes its sparse structure learning capability towards different density levels by exploring all possible parameters across time.

\item The performance of the proposed sparse learning framework is evaluated in the MNIST, Cifar10, Cifar100, and Cifar10-DVS datasets. The experiments show that our proposed structure learning method enables to reduce network parameters efficiently, which facilitates the low power consumption of SNNs during both the training and inference process. 

\end{itemize}

\section{Related Work}

\textbf{Shallow Structures of SNNs.} The early SNNs models with the shallow and memory-saving structure are suitable for rapid data processing when facing a relatively simple problem. For instance, the single-layer Tempotron model manipulates the weight updates according to whether the output neuron fires a spike or not \cite{gutig_tempotron:_2006}. Without considering the concrete firing time of output neurons, the Tempotron model could be trained efficiently to recognize different categories of human gestures collected by the Dynamic vision sensor (DVS) at the millisecond level. However, the Tempotron could only memorize a certain number of spatio-temporal patterns, which are about three times of the network's synapses.
To process and memorize more spatio-temporal patterns, Precise-spike-driven (PSD) synaptic plasticity method takes advantage of the concrete spike timing and employs the error between the actual output spike train and the target spike train to control weight updates \cite{yu2013precise}. The positive errors would trigger long-term potentiation, while the negative errors would contribute to short-term depression. However, these SNNs with relatively shallow structures fail to catch up with the advanced performance of Deep neural networks (DNNs) in complex problems with huge data, such as image recognition and natural language process applications.

\textbf{Deep Structures of SNNs.} Recently, SNNs models with complex and deep structures have been studied to pursue high performance. Due to the non-differential property of the spike train, deep SNNs are hard to be trained directly with Backpropagation (BP) like DNNs \cite{ding2021optimal, xu2021robust}. One of the most commonly used methods is the surrogate gradient, which computes the approximate value gradient by replacing the discrete spike firing process with the given continuous activation function. By the approximated derivative for spike activity with the surrogate gradient, the Spatio-temporal backpropagation (STBP) algorithm combines the layer-by-layer spatial domain and the timing-dependent temporal domain to improve the training efficiency of deep SNNs \cite{wu2018spatio}.
In addition, the local learning rule, such as the Spike-Timing-Dependent Plasticity, which portrays the weight updating between presynaptic and postsynaptic neurons by their chronological relationship of firing times, could also be used to train the deep SNNs in a layer-wise manner. 
Lee et al. propose a two-phase training methodology, which first trains convolutional kernels in an unsupervised layer-specific way, then fine-tunes synaptic weights with spike-based supervised gradient descent backpropagation \cite{lee_training_2018}. This kind of unsupervised local learning solution could help better initialize the parameters in the multi-layer networks prior to supervised optimization and enable convolutional SNNs to be trainable with such deep structures.

\textbf{Structure Pruning of SNNs.} 
Although those above deep SNNs models could achieve high performance comparable to DNNs by extending network structures, they also are perplexed by the enormous trainable parameters like DNNs.
To solve the parameter redundancy problem in deep SNNs, the most commonly used method is parameter pruning after training combined with fine-tuning. In this manner, redundant parameters of trained SNNs could be pruned without losing much performance \cite{deng2021comprehensive}. 

The sparse training of the SNN structure is significant not only because of the strong expressibility comparable to dense training but also due to its potential in the on-chip learning capability once realized on the embedded systems hardware \cite{nguyen2021connection}.

\section{Method}

This section introduces the sparse structure learning approach of the proposed ESL-SNNs. The unified framework of ESL-SNNs is first formally introduced to show the pipeline of the structure learning process. It adapts to different kinds of SNNs, such as multi-layer feedforward SNNs and convolutional SNNs. Then the primary components of the Erdös–Rényi random initialization and connection plasticity evolution are followed to describe the detailed structure learning process. After that, the multi-layer feedforward and convolutional model with the ERL-SNN approach are introduced, respectively.

\subsection{The ESL-SNNs Framework}
As illustrated in Algorithm \ref{SET_framework}, the ESL-SNNs circumvent the parameter redundancy during training by evolving sparse structure dynamically. First, to formulate the completely random topology before training starts, the Erdös–Rényi random graph is employed to initialize the sparse structure of the assigned layer \cite{erdHos1960evolution}. The node in the Erdös–Rényi random graph would be connected to other nodes with the same probability. During training, the weak connections in the sparse layer would be pruned while a certain number of new connections would be generated according to the structure plasticity rule. This evolutionary process is proceeding per $T_{iter}$ iterations, along with the updating of the connection mask. Then the saved connection mask keeps fixed when weight updating, which can constrain the weight matrix when proceeding with the standard network training procedure. The above iterative process is repeated until the end of training.

\textbf{Initialization by the Erdös–Rényi random graph.} Let $H_k$ be the sparse connected layer in SNNs, which contains $n^k$ neurons $[h_1^k, h_2^k, ..., h_{n^k}^k]$. Those neurons in the $H_k$ layer are randomly connected to a certain number of neurons in the $H_{k-1}$ layer. The corresponding weight connections matrix between these two layers is $W \in R^{n^{k-1} * n^k}$. It would be initialized as an Erdös–Rényi random graph, in which the probability of the synaptic connection between the neuron $h_i^k$ in layer $H_k$ and neuron $h_j^{k-1}$ in layer $H_{k-1}$ is defined as:
\begin{equation}
\label{ER_graph}
    p(w_{ij}) = \frac{\epsilon (n^k + n^ {k-1})}{n^k * n^ {k-1}}, 
\end{equation}
where the factor of $\epsilon$ controls the sparsity level of neural connections.

Through the Erdös–Rényi graph initialization, the randomly connected topology of SNNs is applied to the standard training process. However, this random initialization could not guarantee the fast convergence of the defined sparse SNNs towards the given training dataset. Hence, the evolutionary rule should be employed to promote the structure optimization dynamically towards the training dataset during training.

\textbf{Sparse structure evolutionary rule.} Different evolutionary rules could be designed to rewire synaptic connections during training \cite{such2017deep, chenpruning}. To avoid introducing extra parameters to increase memory usage, here, the simple but efficient sparse structure evolutionary rule is employed for effective and fast training. 
There are two steps for connection evolution, including connection pruning and growth. 
Following the pruning rule in \cite{mocanu2018scalable} and the simple magnitude-based pruning rule, the fraction of $\alpha$ weights that are most close to $0$ should be removed. 
To ensure that the pruned network evolves to fit the data, the same fraction of new connections would be regenerated according to the growth rule. Therefore, the connectivity sparsity is stable, and the memory usage maintains at a similar level. Moreover, to ensure the sparse SNNs take full advantage of the information expression capacity of the original large networks, all the connections should be activated during training. Hence, the growing rule would make the connections not activated for a long time to grow and be activated.

Different growing methods are implemented in the sparse structure rewiring procedure of SNNs. 
The gradient-based growth rule forms new synaptic connections by selecting the pruned connections with the largest gradient obtained from the instantaneous weight gradient information in \cite{pmlr-v119-evci20a, DBLP:journals/corr/abs-1907-04840}. 
The weight value of the regenerated synapse connection is initialized to be zero. 
The cosine annealing is employed to decay the proportion of updated connections each time.
\begin{equation}
    f_{decay}(t; \alpha, T_{end}) = \frac{\alpha}{2} (1+ cos(\frac{t\pi}{T_{end}})), 
\end{equation}
where $T_{end}$ is the final iteration to stop updating the sparse connectivity. 
Similarly, the momentum-based growth rule is to regenerate the synaptic connection according to the momentum of parameters \cite{DBLP:journals/corr/abs-1907-04840}.
The random unfired growth rules indicate that the newly generated synaptic connections are controlled randomly and give priority to unfired synapses at the start of the random process. 
In this way, the evolutionary process tends to find a more optimized network structure compared with the fixed sparse structure as the iteration goes on.

\begin{algorithm}[tb]
\caption{The training process of structure learning framework of ESL-SNNs.}
\label{SET_framework}
\textbf{Input Data}: $x_i, i=1, 2, ..., N$.\\
\textbf{Labels of Input Data}:$c_i, i=1, 2, ..., N$. \\
\textbf{Parameters}: The proportion of the updating elements in the mask $\alpha$. The weight mask: $M$. The weight matrix: $W$. The updating iterations: $T_{iter}$.\\

\begin{algorithmic}[1] 
\FOR{each assigned sparse layer of the SNNs}    
\STATE {Initialize the sparse connected layer as the Erd\"{o}s–R\'{e}nyi topology defined in Equation. \ref{ER_graph};}
\ENDFOR
\STATE Initialize training parameters;
\FOR {each training iterations i }   
\STATE Perform standard training procedure;
\STATE Perform weights updates;
\IF {i $\%$ $T_{iter}$ == 0:}
\FOR {each assigned sparse layer of SNNs }  
\STATE (1) Remove a fraction $\alpha$ of synaptic connections by pruning rule.
\STATE (2) Regenerate a fraction $\alpha$ of synaptic connections by growing rule.
\STATE (3) Update the weight matrix of $W$ by element-wise production with weight mask $M$: \\
\STATE W = M $\odot$ W.\\
\ENDFOR
\ENDIF
\ENDFOR
\STATE \textbf{return} The sparse SNNs with $W$.
\end{algorithmic}
\end{algorithm}

\subsection{Multi-Layer ESL-SNNs}

To verify the efficiency of the ESL-SNNs, we first use it to three-layer feedforward SNNs \cite{mostafa_supervised_2018}. The spiking neuron of non-Leaky integrate-and-fire neurons with exponentially decaying synaptic current kernels is employed as the fundamental unit in the network. We assume that there are $N_I$ presynaptic neurons in the Layer $L_I$, and the connection weight between the presynaptic neuron $i$ and the postsynaptic neuron $j$ is $w_{ij}$.  Hence, the membrane potential of neuron $j$ is given by:
\begin{equation}
    \label{LIF_V}
    V_{j}(t) = \sum_{i=1}^{N_I} \Theta(t-t_i) w_{ij}  (1-exp(-(t-t_i))),
\end{equation}
where $t_i$ and $\Theta$ denote the concrete firing time of neuron $i$ and the Heaviside function, respectively. $\Theta$ controls whether the postsynaptic potential contributed by the presynaptic neuron is transmitted to the postsynaptic neuron or not. Only when $t_i<t$, the presynaptic neuron $i$ would transmit the corresponding postsynaptic potential to neuron $j$; otherwise, that potential vanishes. 

The neuron $j$ emits a spike once the value of $V_{j}$ crosses the threshold $V_{thr}=1$. The presynaptic spikes that determine the time point at which the postsynaptic neuron fires the first spike are collected and preserved, defined as the casual set $ C_j = \{ i: t_i < t_j \}$.  Combining with Equation. \ref{LIF_V}, $t_j$ satisfies:
\begin{equation}
\label{t_j_satisfy}
1 = \sum_{i \in C_j} w_{ij} m_{ij}  (1-exp(-(t_j - t_i))).
\end{equation}
Next, the spike times $t_j$ is transformed to z-domain by $exp(t_j) \to z_j$. Hence the first spike of neuron $j$ in the z-domain is given by:
\begin{equation}
\label{z_j_satisfy}
z_j = \frac{\sum_{i \in C_j} w_{ij} m_{ij}   z_i}{\sum_{i \in C_j} w_{ij} m_{ij}   - 1}.
\end{equation}

Then through the above formulations, the output of each neuron, i.e., firing times, could be computed sequentially over feedforward layers. 
In the final output layer, the interpreted cross-entropy loss function in the z-domain is adopted with ensuring that the neuron of the correct class fires earlier than others. 
Assuming the spike time of the output neuron is $z_o$ in the z-domain, and the target class is $g$, the cost of the output layer is given by:
\begin{equation}
\label{cross_entro_loss}
L_{z-domain}(g, z_o) = (-ln \frac{exp(-z_o[g])}{\sum_{k}exp(-z_o[k])}),
\end{equation}
Then the backward propagation proceeds by the gradient descent rule. 


\subsection{Convolutional ESL-SNNs}

The performance of the above three-layer SNNs is limited due to the shallow structure. Hence, convolutional ESL-SNNs are explored in this section. 
The iterative LIF neuron model expressed with the Euler method is adopted to facilitate the information integration and expression in the temporal dimension  \cite{wu2019direct}. According to the membrane potential at $t-1$ and the integrated presynaptic neuron input $I(t)$, the membrane potential $u(t)$ of postsynaptic neuron $j$ is updated as follows:
\begin{equation}
    u(t) = \tau u(t-1) + I(t),
\end{equation}
where $\tau$ denotes the leaky factor and is set to be $0.5$. $ I(t)$ is the product of synaptic weight $W$ and spike input $x(t)$. Similar to the non-LIF model described in the last section, once $u(t)$ crosses the firing threshold $V_{th}$, the neuron fires a spike, and $u(t)$ is set to be $0$. Hence, the neuron output $a(t+1)$ and the membrane updating can be described as:
\begin{equation}
    a(t+1) = \Theta (u(t+1)-V_{th}),
\end{equation}
\begin{equation}
    u(t+1) = u(t+1) (1-a(t+1)),
\end{equation}

To preserve enough information for classification, presynaptic inputs $I(t)$ are integrated with no decay or firing as the output signal for each output neuron. The TET loss function $L_{TET}$ \cite{deng2021temporal} that constrains the output signal at each time point to approach the target distribution is given by:
\begin{equation}
    L_{TET} = \frac{1}{T} \sum_{t=1}^{T} L_{CE} [O(t), y],
\end{equation}
where $T$ is the simulation time length and $L_{CE}$ is the cross-entropy loss function. 
During the training process of each mask update iteration, the weight matrix $W$ is masked by the evolutionary mask $M$. For each element contained in these two matrices, we have $w_{ij} = w_{ij} * m_{ij}$. 

\section{Results}

In this section, we perform a number of experiments to evaluate the effectiveness of the proposed ESL-SNNs. We first introduce the experiment settings including the datasets and model parameters. To verify the scalability of the ESL-SNNs framework over different kinds of SNNs, the evaluations of the ESL-SNNs with three-layer feedforward SNNs and convolutional SNNs are conducted separately on different datasets. The structure sparsity and the structure learning effectiveness are investigated over these two models.

\subsection{The Experiment Settings}

\begin{table*}[!hbt]
  \centering
\begin{tabular}{ccccccc}
\hline
Model    & \begin{tabular}[c]{@{}c@{}}Test Accuracy\\ ($\%$)\end{tabular} & \begin{tabular}[c]{@{}c@{}}Accuracy Loss\\ ($\%$)\end{tabular} & \begin{tabular}[c]{@{}c@{}}Connection \\ Density\end{tabular} & FLOPS  & \begin{tabular}[c]{@{}c@{}}Energy On GPU\\ (J)\end{tabular} & \begin{tabular}[c]{@{}c@{}}Energy On TrueNorth\\ (J)\end{tabular} \\ \hline
SNNs     & 96.7                                                           & -                                                              & 1.0                                                           & 635K & 1.13E-05                                                    & 7.95E-06                                                          \\
ESL-SNNs & 96.58                                                          & -0.12                                                          & 0.16                                                          & 103K & 1.84E-06                                                    & 1.29E-06                                                          \\ \hline
\end{tabular}
\caption{The performance comparison between multi-layer ESL-SNNs and SNNs under the same settings on the MNIST dataset.}
  \label{table_ML_ESLSNN}
\end{table*}

The architecture of the multi-layer feedforward ESL-SNNs is set to be 784-800-10. During training, the learning rate exponentially decays from $0.01$ to $0.0001$, and the batch size is set to $100$. The gradient normalization is applied to avoid the large gradient and small denominator following the design in \cite{mostafa_supervised_2018}. 

\begin{figure}[!t]
	\centering
	\includegraphics[width=1\columnwidth]{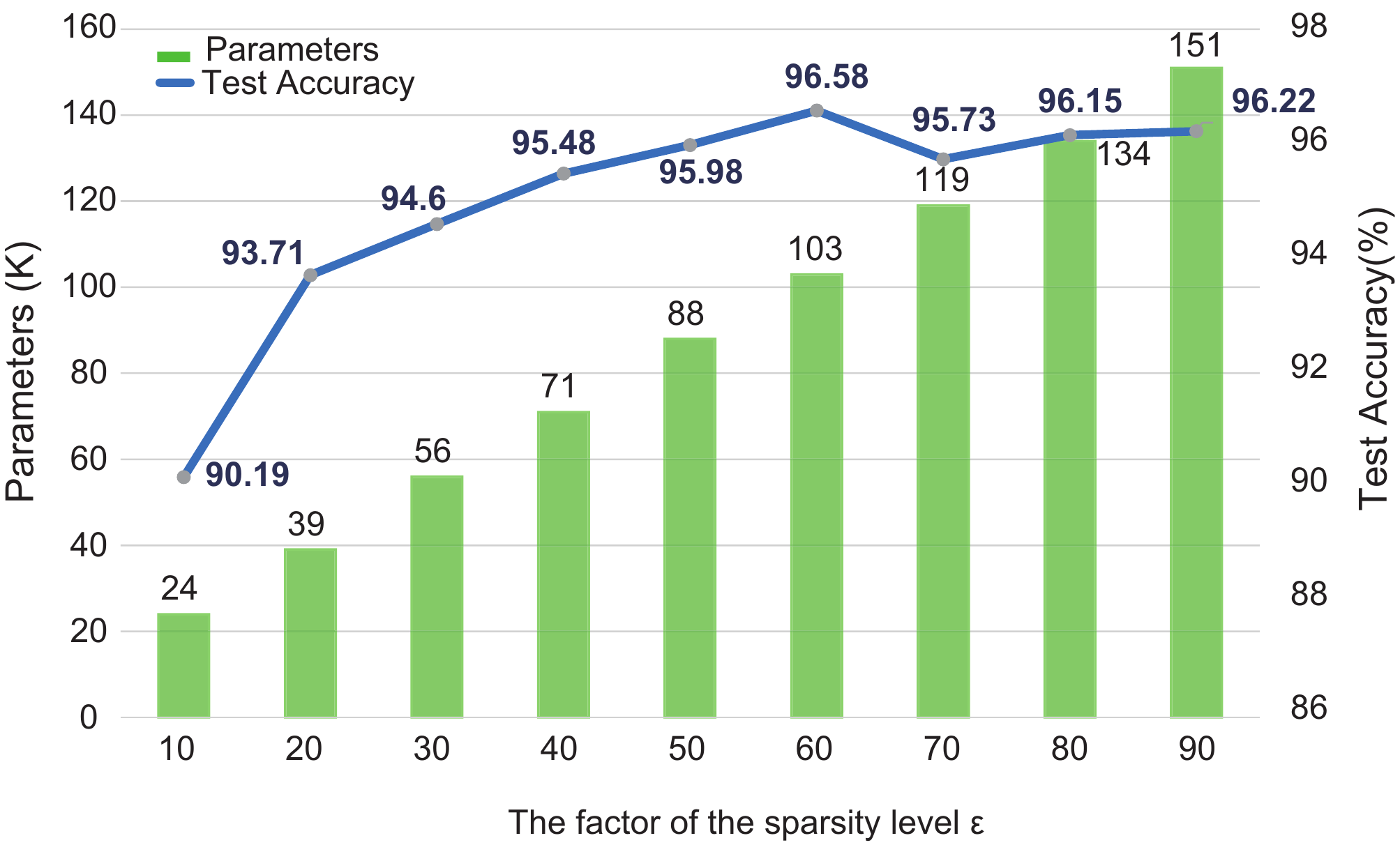}
	\caption{The parameters and test accuracy of multi-layer ESL-SNNs under different sparsity levels (controlled by the factor of $\epsilon$) on the MNIST dataset.}
	\label{fig_epsilon_ML_ESLSNN}
\end{figure}

Convolutional ESL-SNNs are built as VGGSNN (64C3-128C3-AP2-256C3-256C3-AP2-512C3-512C3-AP2-512C3-512C3-AP2) and ResNet19 on DVS-Cifar10, and two Cifar datasets, following the structure in \cite{deng2021temporal}. The simulation time length is set to be $2$ and $4$ to speed up the training speed for Cifar10 and Cifar100. The $T_{iter}$ is set to be $1000$ to control the updating frequency of the weight mask. The learning rate and batch size are $0.001$ and $64$, respectively. The test accuracy is obtained by the model that is saved according to the top-1 accuracy of the validation set within $300$ training epochs. During the training process, the validation set is taken $10\%$ samples randomly from the training set. The average test accuracy over two runs under different random seeds and is reported as the final test accuracy.

\begin{figure*}[!h]
	\centering
	\includegraphics[width=2\columnwidth]{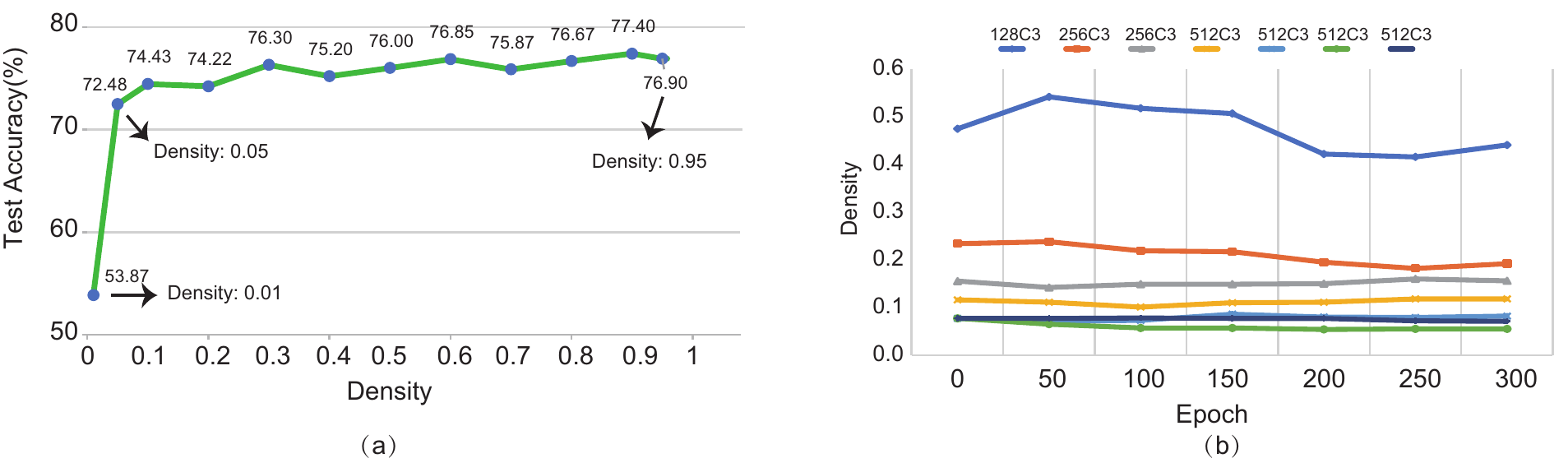}
	\caption{(a) The test accuracy of convolutional ESL-SNNs with the random unfired growth and SET pruning rule under different sparsity levels on Cifar10-DVS. The density in the horizontal axis indicates the percentage of valid connections to total connections. (b) The connection density evolvement of different layers in the sparse VGGSNN.}
	\label{fig_C_ESLSNN}
\end{figure*}

\subsection{Evaluation of Multi-Layer ESL-SNNs}

The performance of multi-layer ESL-SNNs is explored under different sparsity levels. We adjust the sparsity factor $\epsilon$ in Equation. \ref{ER_graph} from $10$ to $90$ to control the sparsity level in the Erdös–Rényi graph initialization. To reduce the influence of the final classification layer, the structure sparsity is only conducted in the connections between the encoding and hidden layers. 
The model parameters and energy consumption are estimated to show the efficiency of ESL-SNNs.

\subsubsection{The performance under different sparsity levels.} The number of parameters in the sparse layer and the test accuracy are analyzed under different sparse levels. As illustrated in Figure \ref{fig_epsilon_ML_ESLSNN}, the heights of the bars denote the number of connections in the sparse layer, and the points on the line chart indicate the test accuracy of multi-layer ESL-SNNs on MNIST. Firstly, with a smaller $\epsilon$, the connections in the sparse layer are sparser. It is consistent with the theory in Equation. \ref{ER_graph}, the connection probability increases as the enlargement of $\epsilon$, hence more connections are preserved. Secondly, when we tune the factor of $\epsilon$ from 10 to 60, the test accuracy improves as the number of connections increases. Since the valuable connections make the network more robust than the fully connected structure. However, when we continue increasing connections, there appears oscillation with a slow recovery when the $\epsilon$ is from 80 to 90. That is because those useless connections may introduce noise to the model.

\subsubsection{Analysis of accuracy and energy efficiency. } The performance of multi-layer ESL-SNNs is compared with the original SNNs under the same settings \cite{mostafa_supervised_2018}. As shown in Table \ref{table_ML_ESLSNN}, multi-layer ESL-SNNs achieve competitive test accuracy with about six times fewer connections. With the total connection amount of $103K$, the test accuracy only drops down $0.12 \%$ compared with the whole fully connected networks with $635K$ connections. Besides, following the total parameter of 103K in the sparse ESL-SNNs, we compare the accuracy of non-sparse fully-connected SNNs with the same parameter amount. That no pruning model achieves about 92$\%$ accuracy, quite lower than the 96.58$\%$ of our method. The result further indicates the effectiveness of the proposed approach. Notably, the above model of our method employs three-layer SNNs in which each neuron is permitted to fire only once. That direct trained single-spike SNNs are quite different from the multi-spike convolutional ESL-SNNs trained by the surrogate gradient method used on Cifar10-DVS. We further evaluate the performance of ESL-SNNs with multi-spike model on MNIST. Our method achieves an accuracy of 97.69$\%$ under 30$\%$ connection density, with no more than 1$\%$ accuracy loss. We adopt these two different kinds of SNNs to illustrate the scalability and generality of our method. 
Therefore, the proposed multi-layer ESL-SNNs have the potential to learn the sparse structure with slight accuracy degradation.

In addition, the energy consumption on GPU and neuromorphic chip are estimated to indicate that the ESL-SNNs can further improve the energy efficiency by virtue of sparse structure. The powers of FLOPS on GPU and SOPS on the neuromorphic chip are obtained from Titan V100 and TrueNorth, respectively. The results of Table \ref{table_ML_ESLSNN} suggest that the proposed ESL-SNNs implemented on neuromorphic hardware platform have about one order of magnitude higher energy efficiency than on GPU platform.

\subsection{Evaluation of Convolutional ESL-SNNs}

Considering the limited performance of multi-layer ESL-SNNs, convolutional ESL-SNNs are employed to explore the potential of ESL-SNNs with the more complex network structure on the larger dataset.

\subsubsection{The performance under different sparsity levels.} 
As illustrated in Figure \ref{fig_C_ESLSNN}(a), the results show that the test accuracies of convolutional ESL-SNNs occur an upward trend along with oscillation as the density of connections enhances. When the connection density distributes from $0.3$ to $0.9$, the test accuracy reduces from $1.18 \%$ to $3.38 \%$ compared with the accuracy of $78.58 \%$ achieved by the model with all connections. Meanwhile, the oscillation phenomenon appears similar to Figure \ref{fig_epsilon_ML_ESLSNN}. This phenomenon suggests that a small number of redundant connections improves the robustness of the network's structure, such as in the situation where the density is $0.3$. 
However, the superfluous redundant connection would seriously drop the accuracy, such as when the density is $0.7$ or $0.95$.

\begin{figure*}[!h]
	\centering
	\includegraphics[width=2\columnwidth]{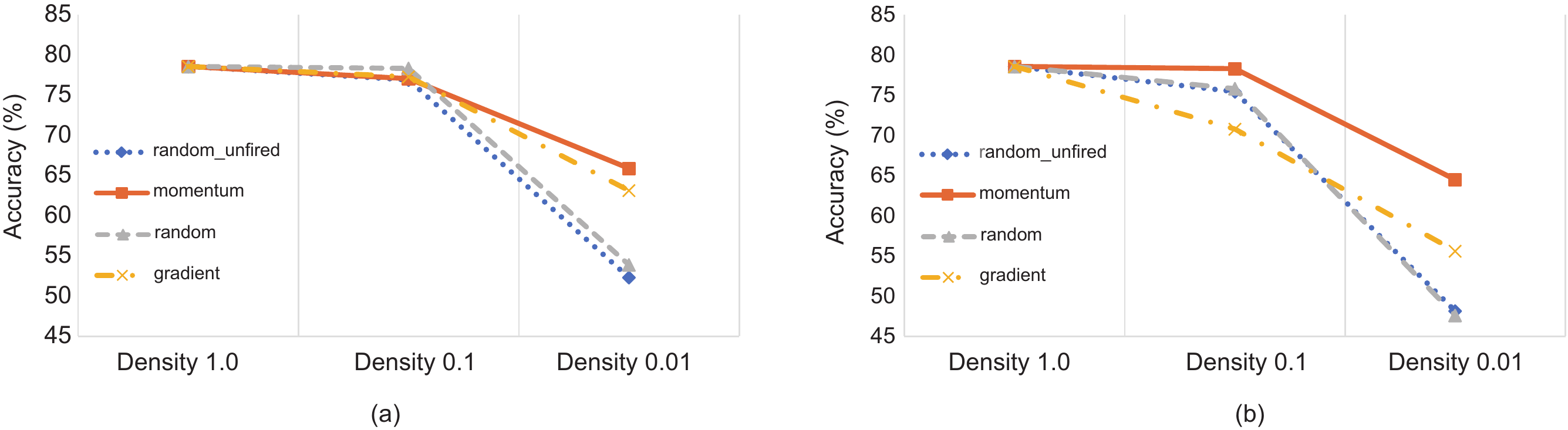}
	\caption{The influence of different evolutionary methods in ESL-SNN. The accuracies on the DVS-Cifar10 dataset are recorded for ESL-SNNs with three different connection densities of $1.0, 0.1$, and $0.01$. (a) The accuracy comparison between different growth rules with the 'magnitude' pruning rule in ESL-SNNs. (b) The accuracy comparison between different growth rules with the 'SET' pruning rule in ESL-SNNs.}
	\label{fig_growth_prun_rule}
\end{figure*}

\begin{table*}[!h]
\centering
\begin{tabular}{cccccc}
\hline
Dataset     & Methods               & \begin{tabular}[c]{@{}c@{}}Network \\ Architecture\end{tabular} & \begin{tabular}[c]{@{}c@{}}Test Accuracy \\ ($\%$)\end{tabular} & \begin{tabular}[c]{@{}c@{}}Accuracy \\ Loss ($\%$)\end{tabular} & \begin{tabular}[c]{@{}c@{}}Connection \\ Density\end{tabular} \\ \hline
            & ADMM-based \cite{deng2021comprehensive}  & 7Conv,2FC                                                       & 89.53                                                           & -3.85                                                           & 0.1                                                           \\
Cifar10          & Grad R \cite{chenpruning}                & 6Conv,2FC                                                       & 92.84                                                           & -0.34                                                           & 0.12                                                          \\ 
    & TET \cite{deng2021temporal}  & ResNet-19                                                       & 92.79                                                           & -                                                               & -                                                             \\
           & ESL-SNNs  & Sparse ResNet-19                                                & \textbf{91.09}                                                  & \textbf{-1.7}                                                   & \textbf{0.5}                                                  \\ \hline
            
Cifar100    & TET \cite{deng2021temporal}                                                          & ResNet-19        & 74.47                           &              -                   & -                             \\
            & ESL-SNNs                                                      & Sparse ResNet-19 & \textbf{73.48}                           & \textbf{-0.99}                           & \textbf{0.5}                           \\ \hline
& Streaming Rollout \cite{kugele2020efficient}     & DenseNet                                                        & 66.8                                                            & -                                                               & -                                                             \\
DVS-Cifar10 & Conv3D \cite{wu2021liaf}    & LIAF-Net                                                        & 71.7                                                            & -                                                               & -                                                             \\
            & TET \cite{deng2021temporal}  & VGGSNN                                                          & 78.58                                                           & -                                                               & -                                                             \\
            & ESL-SNNs & Sparse VGGSNN                                                   & \textbf{78.3}                                                   & \textbf{-0.28}                                                  & \textbf{0.1}                                                  \\ \hline
\end{tabular}
\caption{The performance comparison between convolutional ESL-SNNs and other SNNs models.}
\label{table_C_ESLSNN}
\end{table*}

\subsubsection{The parameters exploring over the training process.}

The connection density evolvement of different layers in the sparse convolutional ESL-SNNs during training is investigated. As shown in Figure \ref{fig_C_ESLSNN}(b), the connection densities of those layers occur the oscillation phenomenon within $200$ epochs for the ESL-SNNs model is evolving to find the optimal structure during training. 
Meanwhile, this connection evolvement tends to be stable after $200$ epochs. 
Moreover, the density level in each layer is beneficial to be proportional to its synaptic connection number. For instance, in the convolutional layer with 128 3*3 kernels, ESL-SNNs are prone to make that layer denser with $47\%$ connections, while tending to make the layer with 512 3*3 kernels sparser with only $8\%$ connections. This phenomenon is consistent with \cite{DBLP:journals/corr/abs-1907-04840}. That is reasonable for the dense layer with more parameters contains more redundancy information compared to the small layer with few parameters, which suggests more parameter space to cut. Hence, the results indicate that the network connections can dynamically prune or regenerate as the training process proceeds.

The influence of different evolutionary methods, containing growth rules and prune rules in ESL-SNN, are analyzed to show the expressibility of sparse training under three network structure density levels from 1.0 to 0.01. As in Figure \ref{fig_growth_prun_rule}, the ESL-SNNs framework with the group of SET prune rule and momentum growth rule is the most stable model under different densities, which has only $ 0.28\%$ and $ 14.08\%$ accuracy loss when the density is 0.1 and 0.01, respectively. It suggests that momentum growth can guide ESL-SNNs to find the most promising synaptic connections during sparse structure updating.
However, the gradient-based growth rule promotes the sparse structure to tend to be similar structures, which in turn limits the expressibility of the SNN model.
Hence, the ESL-SNNs improve the performance of sparse training mainly due to the parameter exploration across the training process.

\subsubsection{Comparison with other state-of-the-art models.}

The performance comparison of the convolutional ESL-SNNs with other state-of-the-art models is investigated to explore the effectiveness on Cifar10, Cifar100 and DVS-Cifar10. As illustrated in Table \ref{table_C_ESLSNN}, ESL-SNNs can achieve competitive performance under quite small connection density with sparse training from scratch. Compared with the TET method with ResNet19 architecture \cite{deng2021temporal}, the ESL-SNNs reach an accuracy of $91.09\%$ and $73.48\%$ when connection density is constrained to $50\%$ with the parameter size of about 6.28M on Cifar10 and Cifar100, respectively. With smaller parameter size compared with the original non-sparse SNNs, the sparse SNNs would consume less energy on computation, memory access, and addressing \cite{lemaire2022analytical}. Although the accuracy of ESL-SNNs falls behind the Grad R model, the advantages of ESL-SNNs over the Grad R model on power saving and memory usage during the whole training procedure should be emphasized. 
Moreover, our ESL-SNNs achieve competitive test accuracy with fewer connections among those state-of-the-art models on DVS-Cifar10. 
With similar or even fewer connections, 
ESL-SNNs have a test accuracy of $78.3\%$, which is higher than the DenseNet with streaming rollout method and the LIAF-Net with the Conv3D components \cite{wu2021liaf}. 
There is only $0.28\%$ (\textless$ 1\%$) accuracy loss for our sparse training with $10\%$ synaptic connections for ESL-SNNs. 
These results suggest the stability of the structure learning of ESL-SNNs on convolutional SNNs with different structures. 
Moreover, it is worth noting that, besides achieving comparable accuracy with sparse structure, the ESL-SNNs have another significant advantage that they could implement the sparse training from scratch, which could reduce memory usage and power consumption and facilitate fast and effective training of large SNNs models.

\section{Conclusion}
The deep structure enables SNNs to catch up with the performance of DNNs in many application scenarios. However, these deep SNNs suffer from parameter redundancy and much memory usage, making them difficult to exploit low power consumption strengths, especially in the training process.
This paper proposes a unified framework of ESL-SNNs to facilitate the sparse training from scratch for SNNs models. The framework of ESL-SNNs can be used for different types of SNNs, such as multi-layer and convolutional SNNs.
It realizes the biologically plausible structure learning procedure inspired by the evolutionary connection rewiring mechanism in the brain. 
The experiment results show that the proposed ESL-SNNs can effectively implement sparse structure learning and achieve competitive accuracy among other SNNs models with fewer parameters. Furthermore, the sparse training of ESL-SNNs improves the expressibility of SNNs with sparse structure, which has the potential for low power consumption, memory usage, and on-chip learning capability when implemented on the embedded hardware.

\newpage

\section{Acknowledgements}

This work was supported by National Key Research and Development Program of China under Grant (No. 2020AAA0105900, No. 2021ZD0109803), National Natural Science Foundation of China under Grant (No. 62236007, No. 62206037), Zhejiang Lab under Grant (No. 2021KC0AC01) and Natural Science Foundation of China (No. U1909202).

\bibliography{ESLSNN}

\end{document}